%% file: root.tex
\documentclass[letterpaper, 10 pt, conference]{source/ieeeconf}  

\IEEEoverridecommandlockouts                              

\usepackage{multirow}
\usepackage{graphicx} 
\usepackage{xspace}
\usepackage[caption=false,font=normalsize,labelfont=sf,textfont=sf]{subfig}

\usepackage{textcomp}
\usepackage{caption}
\usepackage{booktabs}
\usepackage{url}
\usepackage{xcolor}
\usepackage{algorithm}
\usepackage{tabularx}
\usepackage[noend]{algpseudocode} 

\usepackage{amssymb,amsmath,comment,cite}
\usepackage{bm}
\usepackage[colorlinks,linkcolor=blue]{hyperref}
\usepackage{ifthen}
\newboolean{shortver}
\setboolean{shortver}{true}

\title{\LARGE \bf
Catch It! Learning to Catch in Flight with 
Mobile Dexterous Hands
}

\author{Yuanhang Zhang$^{1 \dagger}$, Tianhai Liang$^{2 \dagger}$, Zhenyang Chen$^{4}$, Yanjie Ze$^{5}$ and Huazhe Xu$^{1, 2, 3}$
}

\let\svthefootnote\thefootnote
\newcommand\freefootnote[1]{%
  \let\thefootnote\relax%
  \footnotetext{#1}%
  \let\thefootnote\svthefootnote%
}

\definecolor{darkgreen}{rgb}{0.0, 0.5, 0.0}

\begin{document}
\setlength{\tabcolsep}{6pt} 
\graphicspath{{./source/figures/}}

\maketitle

\thispagestyle{empty}
\pagestyle{plain}
\freefootnote{$\dagger$ Equal Contribution: Yuanhang Zhang led the project, implemented the training in simulation, multi-process controller and real robot deployment; Tianhai Liang conducted the majority of the real robot experiments and contributed to the sim2real and hardware debugging.}
\freefootnote{$^1$Shanghai Qi Zhi Institute. $^2$Tsinghua University, $^3$Shanghai AI Lab. $^4$Georgia Institute of Technology, $^5$Stanford University.}
\freefootnote{Contact:\tt\small{ yuanhanz@andrew.cmu.edu,
huazhe\_xu@mail.tsinghua.edu.cn}.}

\begin{abstract}
    \input{source/abstract}

\end{abstract}

    
    \section{Introduction}\label{dcmm:sec:intro}
    \input{source/intro}

    \input{source/figures/system}
    
    \section{Related Work}\label{dcmm:sec:related}
    \input{source/related}

    \input{source/figures/training}
    
    \section{System Setup}\label{dcmm:sec:system}
    \input{source/system}

    \input{source/figures/controller}
    
    \section{Learning Mobile Dexterous Catching Policies}\label{dcmm:sec:dcmm}
    \input{source/dcmm}
    \section{Experiments}\label{dcmm:sec:experiments}
    \input{source/experiments}

    \section{Conclusion and Limitation}\label{dcmm:sec:conclude}
    \input{source/conclude}

    \newpage

    
    
    \bibliographystyle{plain}
    \bibliography{source/references}


\end{document}

%% file: source/abstract.tex
Catching objects in flight (i.e., thrown objects) is a common daily skill for humans, yet it presents a significant challenge for robots. This task requires a robot with agile and accurate motion, a large spatial workspace, and the ability to interact with diverse objects. In this paper, we build a mobile manipulator composed of a mobile base, a 6-DoF arm, and a 12-DoF dexterous hand to tackle such a challenging task. We propose a two-stage reinforcement learning framework to efficiently train a whole-body-control catching policy for this high-DoF system in simulation. The objects' throwing configurations, shapes, and sizes are randomized during training to enhance policy adaptivity to various trajectories and object characteristics in flight. The results show that our trained policy catches diverse objects with randomly thrown trajectories, at a high success rate of about 80\% in simulation, with a significant improvement over the baselines. The policy trained in simulation can be directly deployed in the real world with onboard sensing and computation, which achieves catching sandbags in various shapes, randomly thrown by humans.
Our project page is available at \href{https://mobile-dex-catch.github.io/}{https://mobile-dex-catch.github.io}.

%% file: source/intro.tex
Humans possess an innate ability to catch thrown objects, a skill that is crucial not only in everyday activities but also in specialized contexts such as athletic sports. The incorporation of similar capabilities in robotic systems has the potential to revolutionize human-robot interaction, particularly in scenarios that involve dynamic handovers. By enabling robots to adeptly perform agile and long-distance catching maneuvers, we can significantly enhance operational efficiency in various applications. Such advancements allow robots to facilitate object transfers between distant locations, thereby completing tasks within the short airborne duration of the objects.

However, existing research on such agile manipulation has notable limitations. Some studies omit mobile platforms~\cite{deguchi2008goal, bauml2010, mirrazavi2016dynamical, kim2014}, restricting the workspace to catch distant objects, while others lack dexterous hands~\cite{dong2020, abeyruwan2023agile}, limiting interaction with diverse objects. In contrast, we develop a mobile manipulator with a dexterous hand, expanding the workspace and enabling adaptability to diverse objects.

There are several challenges to enable a mobile manipulator with a dexterous hand to catch objects in flight: (i)~\textit{accurate and agile whole-body control}: the mobile base and the arm must coordinate to make the arm's end-effector move to the object in flight precisely while the dexterous hand needs to grasp just in time. It also requires agile and real-time movement because the overall execution period only lasts for about 2s, which is the object's flying time in the air. (ii)~\textit{high-dimensional action space}: the system, comprising three components, presents a large action space, which complicates the optimization of the control policy. (iii)~\textit{randomly thrown and diverse objects}: objects are thrown from random positions with random velocities and vary in shapes, which demands a highly adaptive control policy. 

\begin{figure}[t]
\centering
    \includegraphics[width=0.99\linewidth]{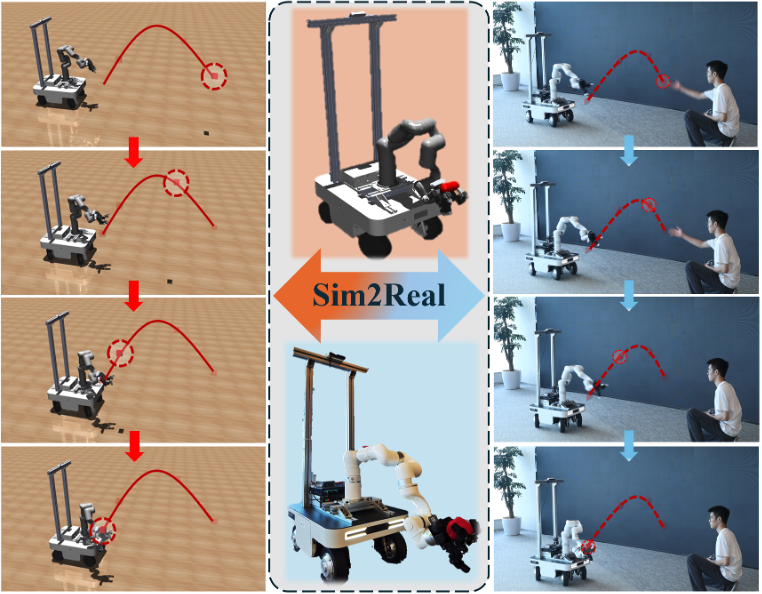}
    \captionof{figure}{\textbf{Sim2Real Illustration of Catching Motions}.}
    \vspace{-0.6cm}
    \label{fig:teaser}
\end{figure}

In this work, we propose \textit{Catch It!}, a learning-based method that leverages reinforcement learning (RL) to learn a whole-body control policy to catch objects in flight in simulation, which can also be used to perform sim-to-real (sim2real) transfer on a real robot.
The key technical contributions of \textit{Catch It!} are summarized as follows:
\begin{enumerate}
    \item \textbf{Whole-Body Control for Mobile Dexterous Catch}:
    We train a unified control policy for the base, arm, and hand to be controlled simultaneously. It enables them to work together seamlessly for coordinated, agile, and accurate objects catching skills.

    \item \textbf{Two-Stage RL Framework}: 
    To deal with the high-dimensional action space, we introduce a model-free RL framework that divides the object-catching task into two subtasks, enhancing training efficiency by focusing on different components in each subtask.

    \item \textbf{Sim2Real for Mobile Dexterous Catch}: 
    We trained the control policy in simulation with careful design to ensure physical and kinematic alignment with the real-world robot. Using sim2real techniques, we successfully deployed our catching policy on the real robot in a zero-shot manner.

    
\end{enumerate}

%% file: source/figures/system.tex
\begin{figure*}[t!]
\centering
    \includegraphics[width=0.98\textwidth]{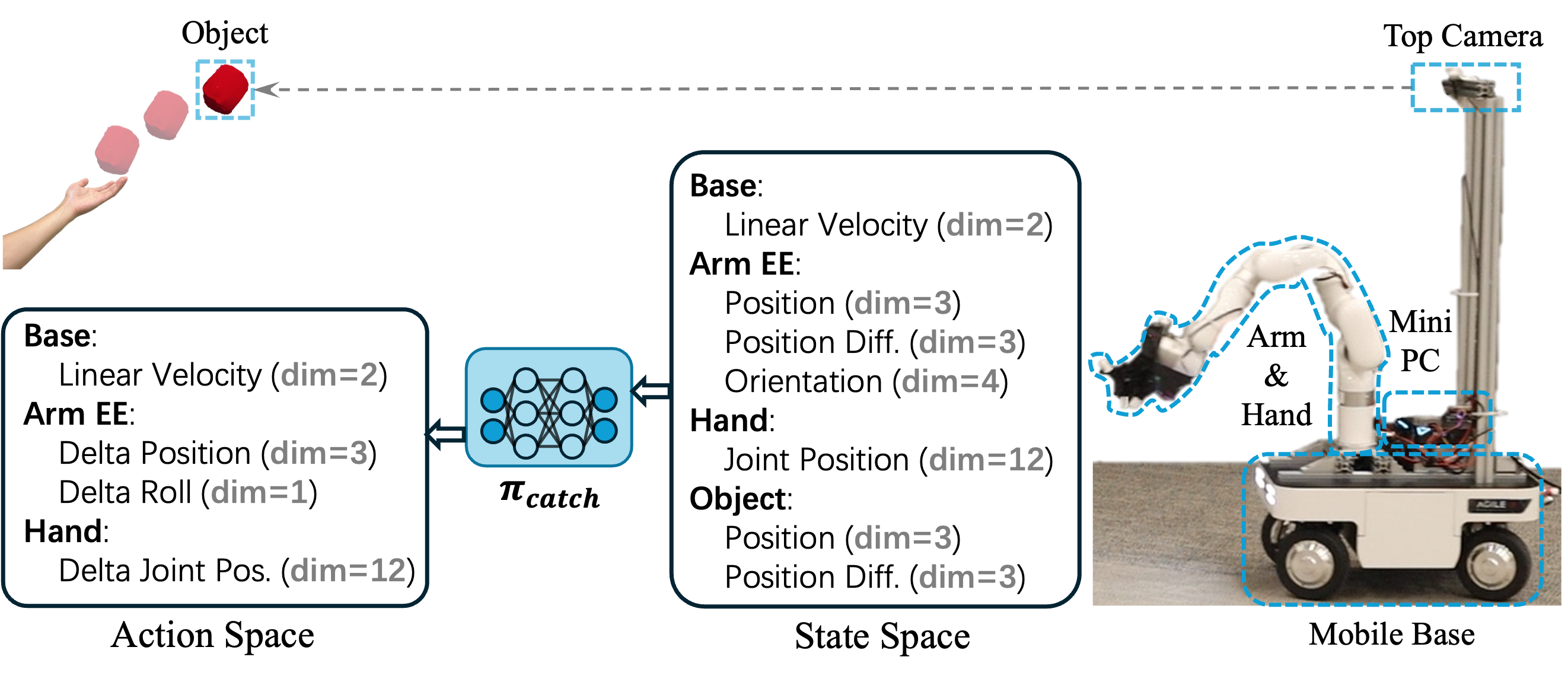}
    \vspace{-0.4cm}
    \captionof{figure}{\textbf{System Overview}. Our system comprises a mobile base, a 6-DoF arm, and a 16-DoF hand, whose goal is to catch objects thrown randomly by humans. The state difference (Diff.) is calculated between two consecutive timesteps. We employ a RGB-D camera to track the object in real-time, and a Mini-PC equipped with a GPU to handle onboard computation.}
    \vspace{-0.6cm}
    \label{fig:real-robot}
\end{figure*}

%% file: source/related.tex
\subsection{Whole-Body Control for Mobile Manipulation}
There has been a large corpus of prior work addressing the whole-body control for mobile manipulation, developing both classical optimization method~\cite{burgess2023architecture, haviland2022holistic, minniti2019whole, 9812280, 9145591, bajracharya2024demonstrating} and learning-based method~\cite{kindle2020whole, sun2022fully, wu2020spatial, yang2023harmonic, lew2023robowip, hu2023causal, fu2024mobile}. In the classical methods~\cite{haviland2022holistic, burgess2023architecture}, a reactive on-the-move architecture is proposed to break down the task (i.e., pick-and-place) into multiple phases. It expresses the coordinately base-and-arm control problem as a quadratic program (QP). Previous works~\cite{9812280, 9145591} also consider motion safety in mobile manipulation and incorporate the collision-free constraints in the Model Predictive Control (MPC). 

While classical methods have achieved some progress in mobile manipulation, they primarily focus on tasks such as navigate-pick-and-place~\cite{9812280, 9145591, bajracharya2024demonstrating} and door-opening~\cite{minniti2019whole}. For more complex dynamic tasks, these methods struggle due to intricate optimization modeling. By contrast, learning-based approaches~\cite{wu2020spatial, lew2023robowip, fu2024mobile, xiong2024adaptive, hu2023causal, burgess2023architecture} offer promising solutions for addressing complex tasks in mobile manipulation. Recent approaches~\cite{fu2024mobile, xiong2024adaptive} leverage RL or Imitation Learning (IL) to enable high-DoF mobile manipulators to perform long-horizon tasks in the real-world scenarios. However, they do not explore the implicit relationships between actions and optimization, which could potentially enhance training efficiency. To address this, Wu \textit{et al}~\cite{wu2020spatial} introduce “spatial~action~maps” for fast learning through pixel-wise action generation, while other work~\cite{hu2023causal} proposes \textit{Causal MoMa}, which uncovers causal dependencies between actions and reward function components to improve training efficiency. Inspired by the existing literature~\cite{hu2023causal, burgess2023architecture}, this paper enhances efficiency by specifically decomposing the object catching task into two subtasks, employing RL with tailored reward designs to train control policies for the key robot components in each subtasks.
 
\subsection{Catching Dynamic Objects with Manipulation}
Catching objects in flight is a key challenge within Dynamic Object Manipulation (DOM). Previous works have explored \textit{slow} DOM using fixed-base manipulators~\cite{salehian2017, marturi2019dynamic, amiranashvili2018motion, huang2023earl, wu2022grasparl, hu2023grasping}.
A prior study~\cite{marturi2019dynamic} develops a learned grasp planner that evaluates potential grasps online for objects moved by humans, selecting from the collision-free and feasible ones. In~\cite{wu2022grasparl, hu2023grasping}, the DOM problem is framed as a two-agent RL task: the robot aims to ``catch'' the object, while the adversarial mover (the object) attempts to evade. This approach enables the trained catching policy to generalize to various object slow-motion patterns.

To learn to catch objects in flight requires greater agility than \textit{slow} DOM, some researchers develop both classical and learning-based methods with an arm, a dexterous hand, and a fixed base~\cite{deguchi2008goal, bauml2010, mirrazavi2016dynamical, kim2014}. Previous study~\cite{bauml2010} formulates the ball-catching problem as a unified nonlinear optimization problem to solve. Another line of work~\cite{mirrazavi2016dynamical} approximates the parameters of linear parameter varying system, which generates a reach-and-follow motion to catch flying objects. Kim \textit{et al}~\cite{kim2014} learns the object falling dynamics and catching policy from throwing demonstrations. Recent work~\cite{huang2023dynamichandoverthrowcatch} proposes a bimanual system that is designed for the throw-and-catch tasks using multi-agent RL.


However, when thrown objects land outside the arm's reach, a mobile platform becomes essential to extend the workspace~\cite{dong2020, abeyruwan2023agile}. For example, prior work~\cite{abeyruwan2023agile} uses a one-dimensional actuator base with MPC and RL strategies to catch thrown objects, while other work~\cite{dong2020} employs an omni-base with a bi-level catching motion planning algorithm and a learning-based controller for tracking. Nevertheless, previous works have not integrated a dexterous hand with mobile manipulators, limiting their generalization to diverse object shapes. In this work, we develop an omni-mobile manipulator with a dexterous hand and use model-free RL to train a catching policy for various randomly thrown objects.

%% file: source/figures/training.tex
\begin{figure*}[ht]
\centering
    \includegraphics[width=0.99\textwidth]{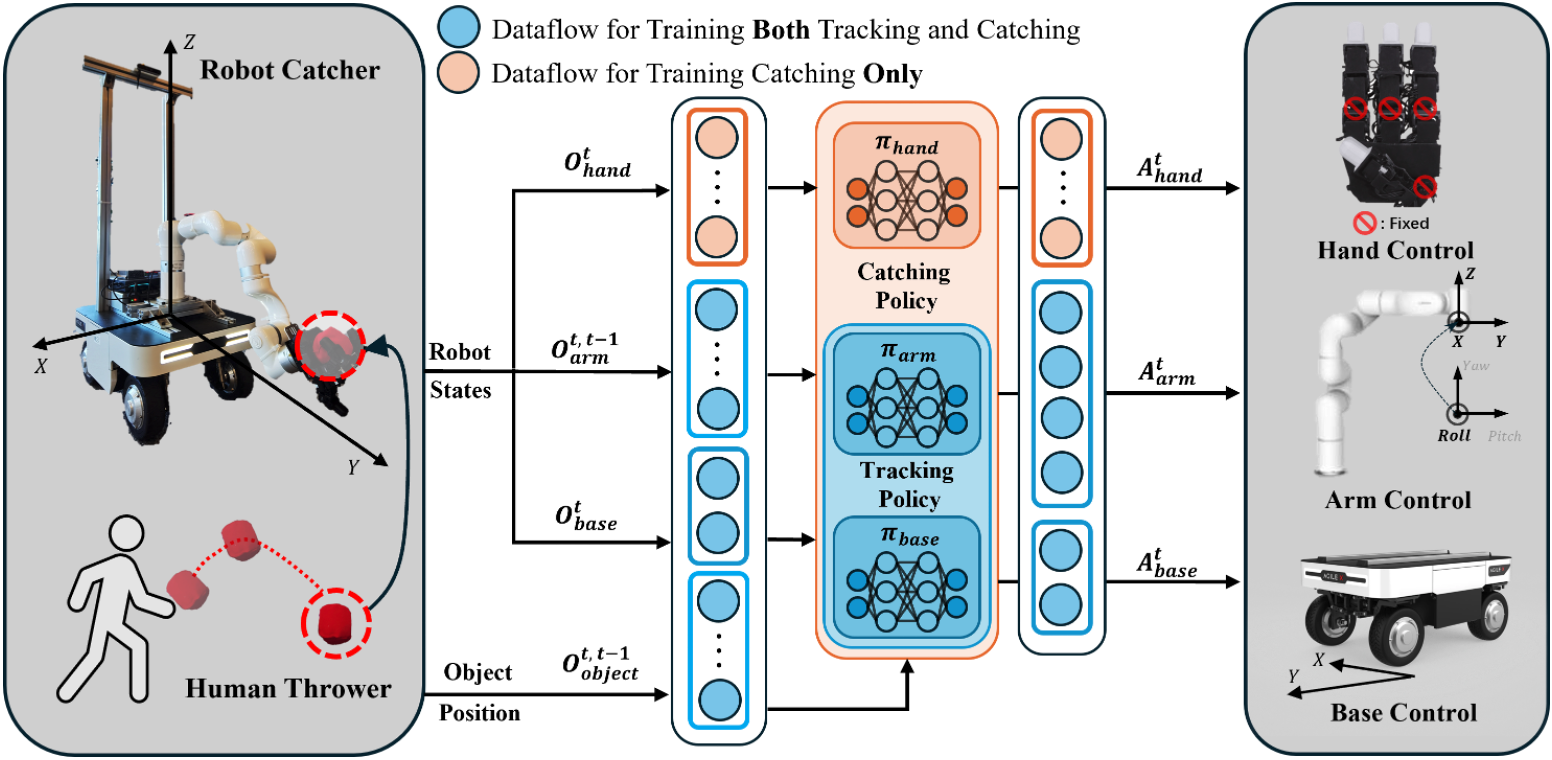}
    \captionof{figure}{\textbf{Two-Stage RL Framework}. Note that we use two consecutive prioproception \bm{$O^{t, t-1}$} as the policy input.}
    \vspace{-0.5cm}
    \label{fig:training}
\end{figure*}

%% file: source/system.tex

\subsection{Task Description}
\label{dcmm:subsec:task}
Our goal is to train a mobile manipulator to catch various objects thrown randomly by humans. Catching objects in flight involves approaching the object with the palm and grasping it stably, which can be divided into two subtasks: (i) The hand needs to track and reach the object. During this phase, only the base and arm are controlled; the hand remains in its initial position. We name this subtask ``tracking task''.
(ii) When the object is about to be reached (i.e., near the palm), the hand needs to grasp the object. Meanwhile, the base and arm are fine-tuned to achieve optimal grasping position. We name this subtask ``catching task''.

\subsection{State and Action Space}
\label{dcmm:subsubsec:saspace}

The state space consists of two consecutive 3D positions of the object and the arm's end-effector, both relative to the arm base fixed on the mobile base, along with the robot’s proprioception, including the base's 2D planar velocity in its body frame and the hand joint positions (Fig.~\ref{fig:real-robot}). We fix 4 hand joints to reduce the space complexity, improving training efficiency while ensuring graspability. During the tracking task, hand states and actions are excluded.


The action space includes the 2D planar velocity of the base, the 12 delta joint positions of the hand, and the 3D delta position as well as the delta roll rotation of the arm's end-effector. We find that controlling the yaw and pitch axes of the arm's end-effector can destabilize the catch policy training, as these movements often lead to unfavorable hand orientations for successful object catching. In contrast, the roll rotation remains beneficial (Sec.~\ref{dcmm:subsubsec:roll}).

\subsection{Real-world Setup}
\label{dcmm:subsec:obj-det}
We construct a mobile manipulator system, as depicted in Fig~\ref{fig:real-robot}. The system consists of a Ranger Mini V2 omni-mobile base, a 6-DoF XArm, and a 12-DoF LEAP Hand. To capture the object's real-time 3D positions in the real world, 
we use an overhead-mounted RealSense D455 camera to extract the object's pixel coordinates and apply a perspective transformation for 3D position estimation relative to the camera. We utilize eye-on-base calibration algorithm to transform this 3D position to the arm's base frame. For the onboard computation, we use a Thunderobot MIX Mini-PC with an i7-13620H CPU and an RTX 4060 GPU. All the components of our robot are powered by the extensive 48V power interface from the Ranger Mini V2.

\subsection{Simulation Setup}
We choose Mujoco~\cite{todorov2012mujoco} as our simulation environment. We use sw2urdf\footnote{\url{https://github.com/ros/solidworks_urdf_exporter}} to build a URDF/MJCF model that mirrors the real robot. For each component of the robot, we develop the PID controller and realize its kinematics respectively:
\subsubsection{Mobile Base}
The motion patterns include \textit{Double-Ackerman}~\cite{hul2020} with forward and yaw velocity as input, and \textit{Parallel} with forward and horizontal velocity as input. Here, we only apply \textit{Parallel} motion pattern, because the base control policy only outputs 2D planar velocity.

\subsubsection{Arm}
We use inverse kinematics (ik) to control the joint positions of the 6-DOF arm, from its end-effector's expected pose. We implement it with two numerical methods: Levenberg-Marquardt (LM) and Quadratic Programme (QP) Method, both of which can find the joint positions corresponding to predefined end-effector position. Besides, the null-space motion is considered to satisfy the angle limits for different joints. For further implementation details, we refer readers to the previous work~\cite{hav2023ik}.

\subsubsection{Hand}
For the LEAP hand, we use a position actuator to control each joint's position. 

%% file: source/figures/controller.tex
\begin{figure*}[ht]
\centering
    \includegraphics[width=0.95\textwidth]{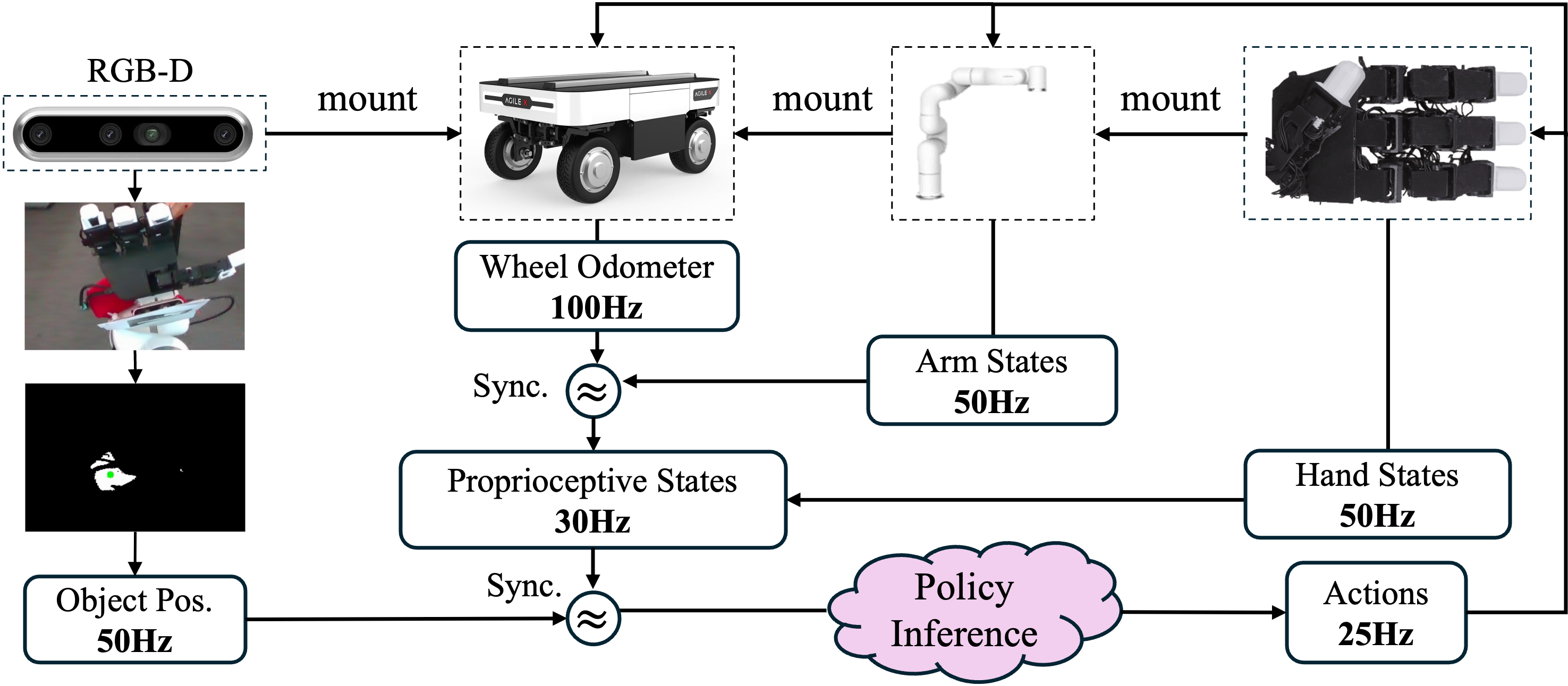}
    \vspace{-0.cm}
    \captionof{figure}{\textbf{Multi-Process Controller}. A ROS-based multi-process controller synchronizes proprioceptive states and object position data for policy inference in real-time control of the mobile manipulator.}
    \vspace{-0.45cm}
    \label{fig:controller}
\end{figure*}

%% file: source/dcmm.tex
\subsection{Two-Stage Reinforcement Learning}
\label{dcmm:subsec:two-stage}
Training the whole-body control policy from scratch to catch objects in flight is inefficient in the face of such complex dynamic task with a high-dimensional action space. Thus, our method \textit{Catch It!} leverages a two-stage reinforcement learning framework to train the catching policy more efficiently. As described in Sec.~\ref{dcmm:subsec:task}, we first train the control policies for the base and arm in the tracking task. Then in the subsequent catching task, we train the hand's control policy while fine-tuning the base and arm's policy from the tracking task, to achieve a better grasping position. In this way, the control policy of the base and arm is pre-trained in the tracking task before starting the catching task, which gives them an initial ability to track and reach the object. Additionally, since the high-dimensional 12-DoF hand movements are unnecessary when the object is distant, fixing the hand in a neutral position during the tracking task of training enhances training efficiency. This two-stage reinforcement training framework is illustrated in Fig.~\ref{fig:training}.

The tracking and the catching task can be formulated as a Partially-Observable Markov Decision Process (POMDP), defined by the tuple $(S, O, A, R, P)$ where $S$ is the state space partially observed by $O$, the observation space, $A$ is the action space, $R: S \times A \mapsto \mathbb{R}$ is the reward function and $P: S \times A \mapsto S$ is the dynamics function. The optimization objective is to learn a parameterized policy $\pi_\theta : O \mapsto A$ that maximizes the expected total episode return, $J(\theta) = E_{\pi=(s_0,a_0,...,s_T)} \sum_{t=0}^T r(s_t, \pi_\theta(o_t))$. The state space consists of the object positions,  which are approximated from the raw observations generated by the real-time object tracking system described in Sec.~\ref{dcmm:subsec:obj-det}, thereby justifying the POMDP categorization. In this work, we use Proximal Policy Optimization (PPO)~\cite{schulman2017ppo} to train our neural network.

\subsection{Reward Design}
Careful reward design in RL is the key to train a robust policy successfully. In both tasks, we reward the policy approaching the object and the orientation alignment between the palm and object. We also give a high reward for the the palm touching the object. Finally, we discourage excessive motion via penalizing policy output, and joint limit violation. 

Given the times $t$, the object's 3D position $\boldsymbol{p_t}$ and velocity $\boldsymbol{v_t}$ (estimated as the difference between consecutive positions), the end-effector's position $\boldsymbol{e_t}$, the $z$-axis vector $\boldsymbol{\hat{u_t}}$, the previous closest hand-to-object distance $\boldsymbol{d_{t-1}}$, and the policy output $\boldsymbol{a_t}$, the detailed reward definitions are:
\begin{itemize}
    \item \textbf{Object Position Reward}~(track/catch): The difference of hand-to-object distance in two consecutive time steps during the episode:~$r_t^{pos} = \lVert \boldsymbol{d_{t-1}} \rVert_2 - \lVert \boldsymbol{e_t} - \boldsymbol{p_t} \rVert_2$.
    \item \textbf{Object Precision Reward}~(track/catch): This reward scales the $d_t$ with an exponential function, which facilitates learning the policy to approach the target with a higher precision~\cite{mah2018}:~$r_t^{pre} = \exp(-50 \cdot \lVert \boldsymbol{d_t} \rVert_2^2)$.
    \item \textbf{Object Orientation Reward}~(track/catch): This reward is computed as the dot product of the delta position vector of the object and the $z$-axis of the palm, clamped between -1 to 1:~$r_t^{orient} = clamp(\boldsymbol{v_t} \cdot \boldsymbol{\hat{u_t}}, -1, 1)$.
    \item \textbf{Object Touch Reward}~(track): A binary reward is given if the palm touches the object: $r_t^{touch}=1~or~0$.
    \item \textbf{Object Stability Reward}~(catch): This reward is computed according to the time length when the object is held by the hand:~$r_t^{stab}=\Delta t_{grasp}$.
    \item \textbf{Control Penalty}~(track/catch): To avoid excessive motion, we penalize the policy output: $r_t^{ctrl}=\lVert \boldsymbol{a_t} \rVert^2_2$.
    \item \textbf{Constraint Penalty}~(track/catch): A binary penalty is provided if the robot joints exceed their joint limits:~$r_t^{cstr}=-1~or~0$.
\end{itemize}
The final reward at $t$, is computed as the weighted sum of the previously mentioned reward terms, each multiplied by a respective scaling coefficient $l_k$:~$r_t^{track/catch} = \sum l_k \cdot r_t^k$.


%% file: source/experiments.tex
\subsection{Experiment Settings}
\subsubsection{Training Settings}
We create 64 parallel environments to collect large amounts of trajectories of various objects to train our tracking and catching policy. The training process is performed on 128 Intel i7 CPUs and a single NVIDIA GeForce RTX 3090 Ti GPU.

\subsubsection{Frequency Settings}
In the simulation, the basic step frequency is set to 500 Hz. The policy inference is made every 20 simulation steps, corresponding to a frequency of 25 Hz. During each set of 20 simulation steps, the robot consistently executes the same action as determined by the latest inference. When deploying the control policy on the real robot, we maintain a control frequency of 25 Hz, ensuring consistency with the simulation.
\subsubsection{Thrown Object Settings}
\label{dcmm:subsec:object}
\begin{figure}[tb]
    \centering
    \includegraphics[width=1\linewidth]{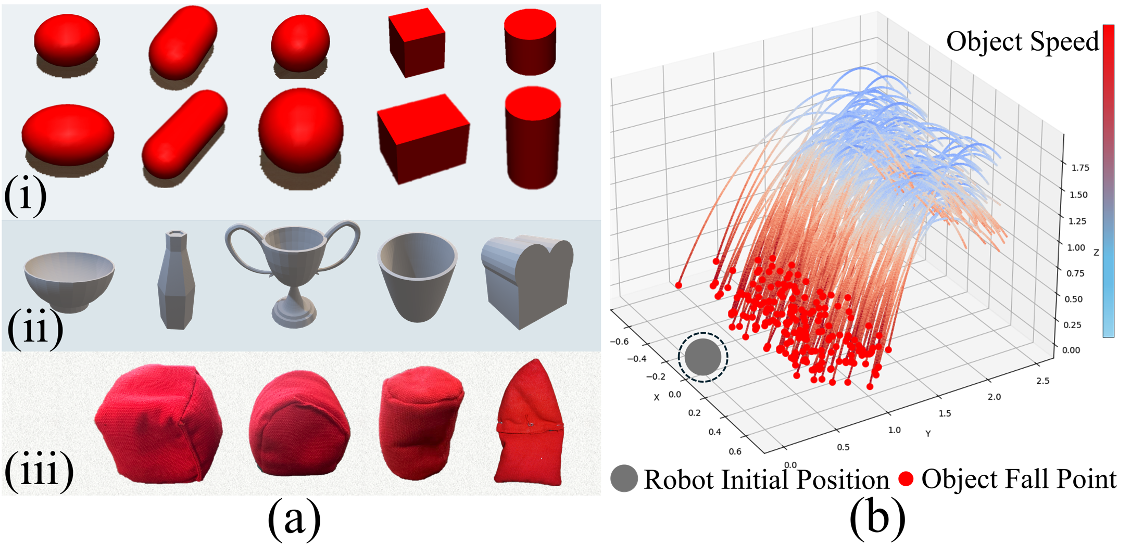}
    \vspace{-0.6cm}
    \caption{\textbf{Object Set Overview}. (a): (i) Objects in training; (ii) Objects in evaluation; (iii) Objects in real-world deployment. (b) Random Throwing Trajectory Visualization.}
    \label{fig:objs}
    \vspace{-0.4cm}
\end{figure}

In the training process, We use objects in 5 different shapes: a box, a sphere, an ellipsoid, a cylinder and a capsule, as shown in Fig.~\ref{fig:objs}~(a)~(i). Among these objects, the cylinder and capsule may require the robot to utilize all of its fingers~\cite{huang2023dynamichandoverthrowcatch}. We randomize the object's shapes, sizes, mass, and damping in each environment initialization only. To validate the policy's generalization in simulation, we test with 5 unseen object shapes, as shown in Fig.~\ref{fig:objs}~(a)~(ii). In the real-world experiments, we employ sandbags in 4 different shapes, as shown in Fig.~\ref{fig:objs}~(a)~(iii): a ball, a cube, a cylinder, and an irregular combination formed by stitching together a triangular sandbag and a flat sandbag.

To collect diverse object trajectories during training, we randomize the initial positions and velocities of the obejcts in each episode. As shown in Fig~\ref{fig:objs}~(b), the farthest landing point is approximately 1.5m from the robot's starting position, which is beyond the arm's reach (about 0.8m), necessitating the use of a mobile base.
\subsubsection{Baselines}
We compare our two-stage reinforcement learning framework with the following two baselines:
\begin{itemize}
    \item \textbf{One-Stage without Tracking Task}: In the one-stage baseline, we skip the tracking task and directly train the catching task from scratch. The base and arm's control policy would not be pre-trained from the tracking task.
    
    \item \textbf{Two-Stage without Arm's Roll}: According to Sec.~\ref{dcmm:subsubsec:saspace}, we remove the rolling action of the arm but still train in a two-stage manner.
\end{itemize}

\subsection{Simulation Results}
We first compare our two-stage training method with the two baselines on their training performance in simulation. Then, we evaluate their success rate in simulation with the 8 unseen objects. In the tracking task, if the palm touches the objects in flight, we consider it as a tracking success. In the catching task, if the object keeps being held in hand until the episode's maximum time, which is set to 2.5s, we consider it a catching success.

\begin{figure}[tb]
    \centering
    \includegraphics[width=1.0\linewidth]{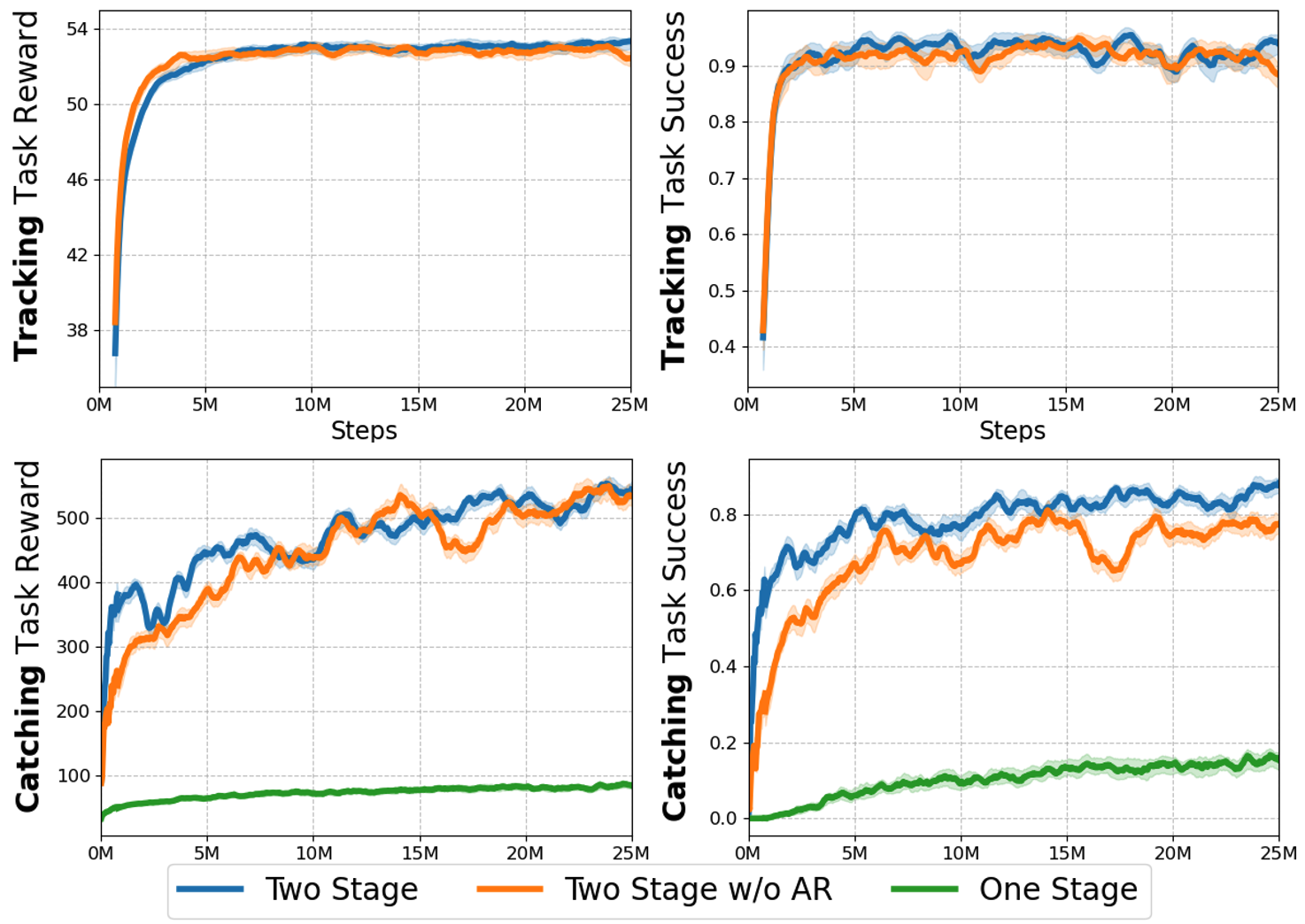}
    \caption{\textbf{Training Curves}. The \textcolor{blue}{blue}, \textcolor{orange}{orange}, and \textcolor{darkgreen}{green} curves represent the two-stage (T.S.), two-stage without arm’s roll (T.S. w/o AR), and one-stage (O.S.) methods, respectively. The first row corresponds to the episode rewards and success rates for the tracking task over the training steps, while the second row shows the same metrics for the catching task.}
    \label{fig:curves}
    \vspace{-0.6cm}
\end{figure}

\subsubsection{Two-Stage versus One-Stage}
Fig.~\ref{fig:curves} shows that the two-stage method (ours) significantly outperforms the one-stage in both training efficiency and final success rate, highlighting both the superiority and necessity of our method for achieving great catching performance discussed in Sec.~\ref{dcmm:subsec:two-stage}.

\subsubsection{Two-Stage versus Two-Stage w/o AR}
\label{dcmm:subsubsec:roll}
Fig.~\ref{fig:curves} shows that while ours and two-stage w/o arm rolling both achieve efficient training and high success rates in the tracking task, our method stands out in terms of success rates in the catching task. This advantage is due to the roll action, which benefits the orientation reward and optimizes the palm's orientation for catching.

\subsubsection{Evaluation}
As shown in Table~\ref{tab:eval}, our method outperforms the other two baselines in the catching success rate for unseen objects. Moreover, our trained policy demonstrates robust generalization to unseen objects, achieving high catching success rates across various shapes not encountered during training. This indicates the effectiveness and adaptability of our method, making it suitable for deployment on real-world robots with unseen object geometries.

\begin{table}
\centering
\begin{tabular}{lccccc}
    \toprule
    Track S.R. (\%) & Bowls & Bottles & Win-Cups & Cups & Breads\\
    \midrule
    T.S. w/o A.R. & 88\textpm 4 & 92\textpm 3 & 90\textpm 5 & 92\textpm 5 & 91\textpm 4\\
    
    \midrule
    T.S. (ours) & \textbf{92\textpm 3} & \textbf{90\textpm 4} & \textbf{88\textpm 3} & \textbf{94\textpm 5} & \textbf{95\textpm 4}\\
    \bottomrule
    
    \toprule
    Catch S.R. (\%) & Bowls & Bottles & Win-Cups & Cups & Breads\\
    \midrule
    O.S. & 22\textpm 2 & 10\textpm 3 & 6\textpm 2 & 13\textpm 2 & 15\textpm 3\\
    \midrule
    T.S. w/o A.R. & 80\textpm 6 & 73\textpm 4 & 61\textpm 7 & 77\textpm 5 & 75\textpm 3\\
    
    \midrule
    T.S. (ours) & \textbf{84\textpm 5} & \textbf{78\textpm 6} & \textbf{65\textpm 3} & \textbf{80\textpm 4} & \textbf{80\textpm 3}\\
    \bottomrule
\end{tabular}
\caption{\textbf{Evaluation of Unseen Objects in Simulation}. It shows the average Success Rate (S.R.) in the tracking and catching tasks for 200$\times$64 trials.}
\vspace{-0.2cm}
\label{tab:eval}
\end{table}

\begin{table}
\centering
\begin{tabular}{lcccc}
    \toprule
    Track S.R. (\%) & Cube & Sphere & Cylinder & The Irregular \\
    \midrule
    T.S. w/o LPF & 10 & 10 & 5 & 15\\
    
    \midrule
    T.S. (ours) & \textbf{70} & \textbf{65} & \textbf{70} & \textbf{75}\\
    \bottomrule
    
    \toprule
    Catch S.R. (\%) & Cube & Sphere & Cylinder & The Irregular \\
    \midrule
    T.S. w/o LPF & 0 & 0 & 0 & 5\\
    
    \midrule
    T.S. (ours) & \textbf{25} & \textbf{25} & \textbf{15} & \textbf{20}\\
    \bottomrule
\end{tabular}
\caption{\textbf{Evaluation in Real Robot Deployment}. It shows the average Success Rate (S.R.) in the tracking and catching tasks for 40$\times$4 trials.}
\vspace{-0.6cm}
\label{tab:eval_real}
\end{table}

\subsection{Sim2Real Transfer}
There remains a large sim2real gap due to the complexity of our mobile manipulator system. To bridge the sim2real gap as much as possible, we leverage the following techniques:
\begin{itemize}
    \item \textbf{Low-Pass Filter}: For the Ranger Mini V2 mobile base, we observe that, unlike in simulation where the base can steer immediately while maintaining forward motion, significant adjustments to the steering angle during forward motion are not feasible in the real world. This limitation causes the base to brake with frequent, large steering changes in velocity commands. To mitigate this, we applied a Low-Pass Filter (LPF)~\cite{lowpass} to smooth the velocity commands, as shown in Fig.~\ref{fig:lpf}, ensuring they are executable by the mobile base in the real world.

    \begin{figure}[t]
        \centering
        \includegraphics[width=1.0\linewidth]{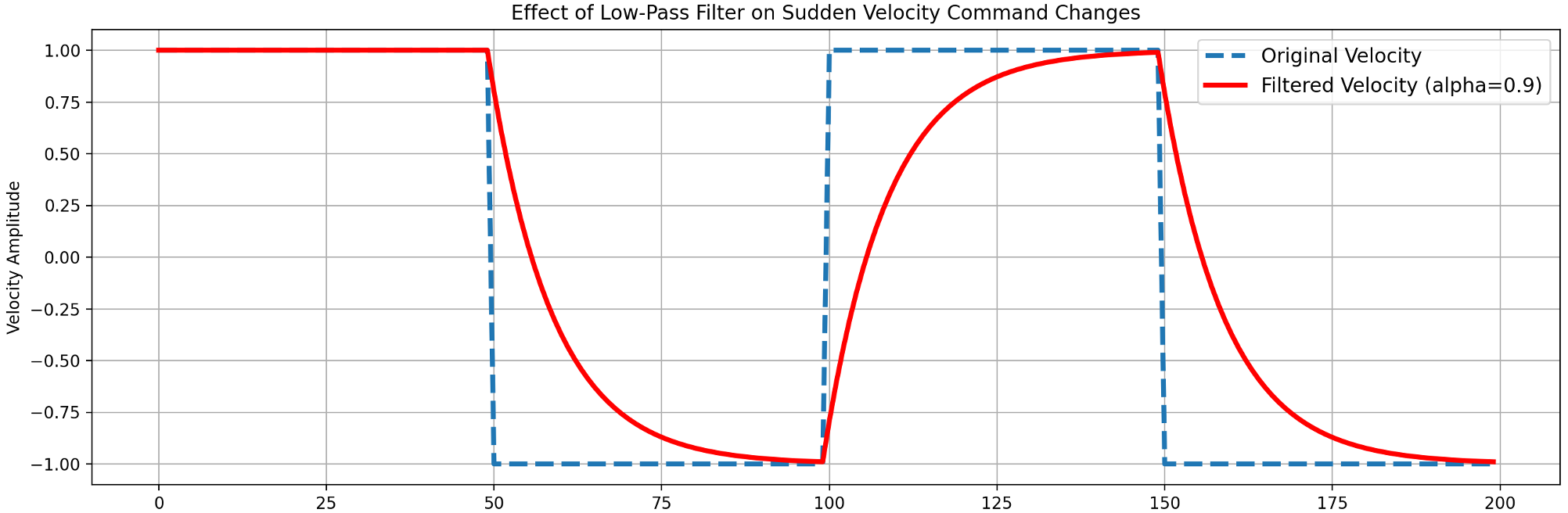}
        \vspace{-0.4cm}
        \caption{\textbf{Low-Pass Filter Curve}. The comparison between the original and filtered velocity commands, using a recursive coefficient of 0.9, consistent with the real robot deployment.}
        \label{fig:lpf}
        \vspace{-0.2cm}
    \end{figure}

    \begin{figure}[t]
        \centering
        \includegraphics[width=1.0\linewidth]{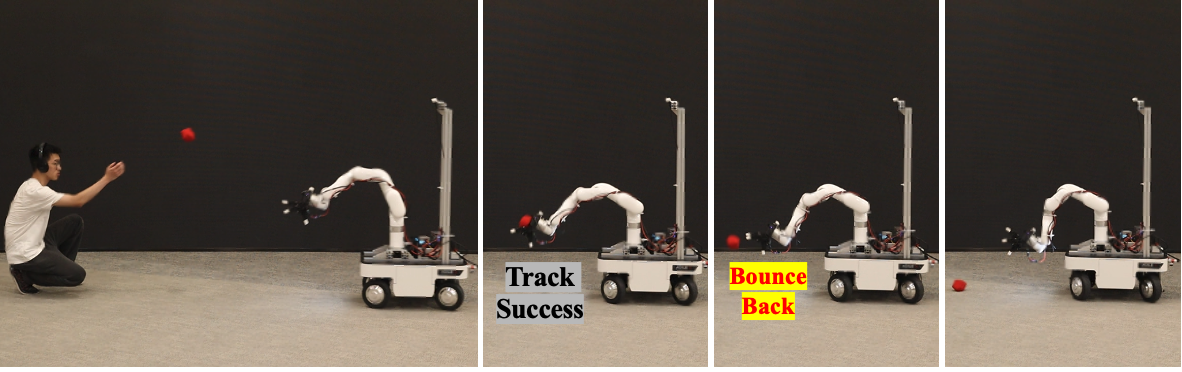}
        \vspace{-0.4cm}
        \caption{\textbf{Failure Case}. The object bounces off the palm.}
        \label{fig:fail}
        \vspace{-0.6cm}
    \end{figure}
        
    \item \textbf{System Identification}: We employ system identification to align the behavior of the PID controllers for the base, arm, and hand between the simulation and the real world. This process serves as a preliminary estimation of the PID parameters within the simulation, which subsequently facilitates the domain randomization process.
    \item \textbf{Domain Randomization}: Domain Randomization has been proven to be effective in the sim2real transfer~\cite{dr2017}. In addition to the randomization of thrown objects discussed in Sec.~\ref{dcmm:subsec:object}, we apply Domain Randomization to the PID parameters, the gravity, the timing of throwing objects, the observation noise, and the action noise. It is important to note that randomizing the throw timing is crucial, as in real-world scenarios, humans typically throw objects at unpredictable moments.
\end{itemize}

\subsection{Real Robot Deployment}
\subsubsection{Controller}
We develop a multi-processing control system based on ROS to manage the various components of the mobile manipulator, as depicted in Fig.~\ref{fig:controller}. Proprioceptive states from the base, arm, and hand are collected at different frequencies and synchronized to 30 Hz. The object’s position is obtained from an RGB-D camera operating at 50 Hz. Both the proprioceptive data and the object’s positional information are synchronized and used as inputs for inferring our whole-body control policy, which runs at 25 Hz and matches the control frequency in simulation.
\subsubsection{Deployment Result}
We deployed the trained catching policy on the real robot across 160 trials, with 40 trials per object shape. Each set of 40 trials included 20 with the LPF and 20 without. As shown in Table~\ref{tab:eval_real}, the success rates for both tracking and catching were low without LPF. In contrast, with LPF, we achieved a high tracking success rate of approximately 70\%, demonstrating the effectiveness of LPF and the robustness of the whole-body control policy trained in simulation. The policy also successfully caught objects of all shapes, highlighting its adativeness in real-world scenarios with varied object geometries. However, the catching success rate did not exceed 25\%, primarily due to the elasticity of the objects (as shown in Fig~\ref{fig:fail}), which introduced challenges not present in the simulation. Additionally, the RGB-D camera used for position tracking generated errors when the object moved quickly or was occluded by the hand. We believe integrating a global localization system, such as VICON, could improve catching performance by providing more accurate and robust object tracking.

%% file: source/conclude.tex

In this work, We present \textit{Catch It!}, addressing the challenge of catching flying objects with a mobile manipulator and dexterous hand. We propose a two-stage reinforcement learning framework which efficiently trains a whole-body control policy to catch randomly thrown objects in various shapes. Results show high success rates in simulation and successful real-world deployment using sim2real techniques.

However, when deploying the trained policy in the real world, our system may struggle with elastic objects, which tend to rebound off contact surfaces, particularly when both of the object and robot move at high speeds. Therefore, in the future, we plan to refine the hand's grasping strategy with tactile sensors to better manage rebounding dynamics. We are committed to releasing the simulation training code and providing references for real robot deployment.